\documentclass{article}

\PassOptionsToPackage{numbers}{natbib}

\usepackage[preprint]{neurips_2022}

\usepackage[utf8]{inputenc} 
\usepackage[T1]{fontenc}    
\usepackage{hyperref}       
\usepackage{url}            
\usepackage{booktabs}       
\usepackage{amsfonts}       
\usepackage{nicefrac}       
\usepackage{microtype}      
\usepackage{xcolor}         
\usepackage{graphicx}
\usepackage{algorithm}
\usepackage{algpseudocode}

\title{EXPANSE: A Deep Continual / Progressive Learning System for Deep Transfer Learning}

\author{%
  Mohammadreza Iman\\
  Department of Computer Science\\
  Franklin College of Arts and Sciences\\
  University of Georgia\\
  Athens, GA, USA \\
  \texttt{m.iman@uga.edu} \\
  \And
  John A. Miller\\
  Department of Computer Science\\ 
  Franklin College of Arts and Sciences\\
  University of Georgia\\
  Athens, GA, USA\\
  \texttt{jamill@uga.edu} \\
  \And
  Khaled Rasheed\\
  Institute for Artificial Intelligence\\
  Franklin College of Arts and Sciences\\
  University of Georgia\\
  Athens, GA, USA\\
  \texttt{khaled@uga.edu}\\
  \And
  Robert M. Branch\\
  Learning, Design, and Technology\\ 
  Mary Frances Early College of Education\\
  University of Georgia\\ 
  Athens, GA, USA\\
  \texttt{rbranch@uga.edu}
  \And
  Hamid R. Arabnia\\
  Department of Computer Science\\ 
  Franklin College of Arts and Sciences\\
  University of Georgia\\
  Athens, GA, USA\\
  \texttt{hra@uga.edu}
}

\begin{document}

\maketitle

\begin{abstract}
Deep transfer learning (DTL) techniques try to tackle the limitations of deep learning, the dependency on extensive training data and the training costs, by reusing obtained knowledge from source data for target data. However, the current DTL techniques suffer from either catastrophic forgetting dilemma (losing the previously obtained knowledge) or overly biased pre-trained models (harder to adapt to target data) in finetuning pre-trained models or freezing a part of the pre-trained model, respectively. Progressive learning, a sub-category of DTL, reduces the effect of the overly biased model in the case of freezing earlier layers by adding a new layer to the end of a frozen pre-trained model. Even though it has been successful in many cases, it cannot yet handle distant source and target data. We propose a new continual/progressive learning approach for deep transfer learning to tackle these limitations. To avoid both catastrophic forgetting dilemma and overly biased-model problem, we expand the pre-trained model by expanding pre-trained layers (adding new nodes to each layer) in the model instead of only adding new layers. Hence the method is named EXPANSE. Our experimental results confirm that we can tackle distant source and target data using this technique. At the same time, the final model is still valid on the source data, achieving a promising deep continual learning approach. Moreover, we offer a new way of training deep learning models inspired by the human education system. We termed this two-step training: learning basics first, then adding complexities and uncertainties. The evaluation implies that the two-step training extracts more meaningful features and a finer basin on the error surface since it can achieve better accuracy in comparison to regular training. EXPANSE (model expansion and two-step training) is a systematic continual learning approach applicable to different problems and DL models.
\end{abstract}



\section{Introduction}

Since the development of traditional machine learning algorithms in the 1980s, Transfer Learning has been a thrilling field of research for scientists. Transfer Learning (TL) is about using obtained knowledge from a source domain/data to facilitate learning on a target domain/data, also known as Domain Adaptation \cite{ef}. Traditional ML models are less dependent on extensive training data; the training process is more straightforward than Deep Learning (DL) models. Therefore, transfer learning in deep learning, known as Deep Transfer Learning (DTL) \cite{e2}, is much more in demand and has been used vastly in the last decade to deal with DL's two aforementioned significant constraints

\begin{figure}[h]
  \centering
\includegraphics[width=0.8\textwidth]{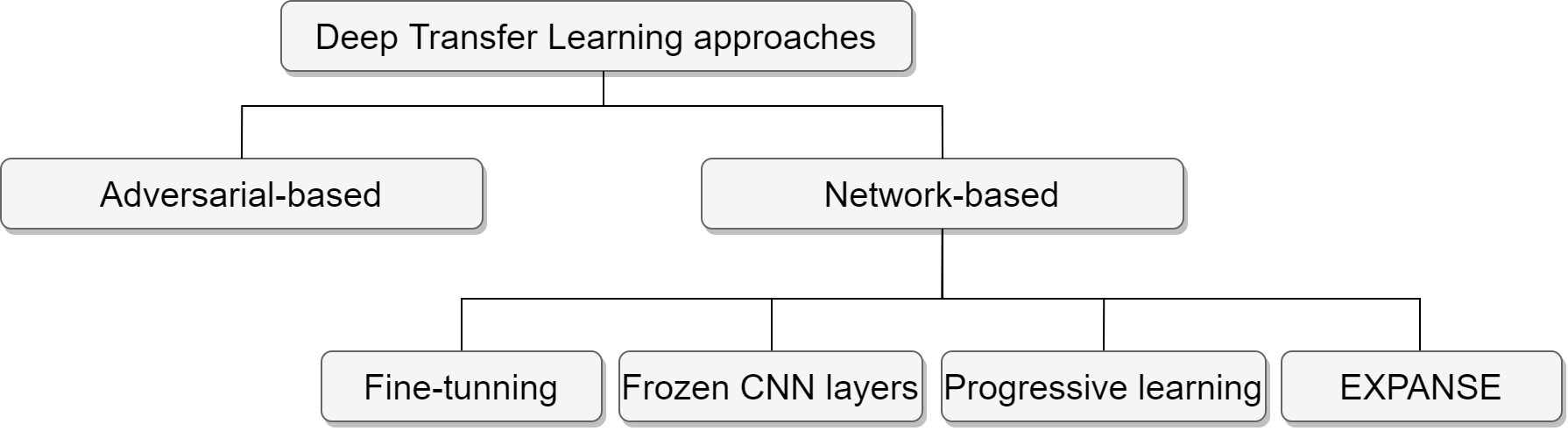}
  \caption{Deep transfer learning common approaches.}\label{fig1}
\end{figure}

Deep Transfer Learning (DTL) methods are mostly Network/Model-based: (i) finetuning a pre-trained model; (ii) freezing a part of a pre-trained model and finetuning the rest; and (iii) progressive learning. Another approach in DTL is Adversarial-based, which is not as common as model-based approaches \cite{e2}. Our proposed methodology, EXPANSE, is also a network/model-based approach. Figure \ref{fig1} shows the hierarchy of common approaches in DTLs and the EXPANSE. For in-depth information on deep transfer learning, please refer to \cite{e2}.

Depending on the approach from finetuning to frozen layers, DTL suffers from catastrophic forgetting or an overly biased model. The catastrophic forgetting dilemma happens while a pre-trained model trains on target data; in most cases, the previously obtained knowledge is partially or entirely wiped out \cite{ec,ec2} . The other end of this spectrum is when several layers of a pre-trained model are frozen to train on target data. In this case, the pre-trained model is overly biased on source data and will not adapt to target data. A consequence of these two constraints is that we can not reach continual learning through current DTLs. In continual learning, a model should learn new skills while still handling the old skill(s).

Progressive learning / Progressive Neural Network (PNN) \cite{eg,epp} was introduced by the Google Deep Mind project in 2016, which is the closest attempt to mimic the human ability of continual learning (building a new skill on top of a previously learned skill). PNN has been applied to various problems, such as natural language processing and image-related tasks, successfully since 2016 \cite{ep1,ep2,ep3,ep4}. In PNN, they freeze the pre-trained model on source data and train on the target data; they add a new layer(s) to the end of the model (near the output). Still, the method is overly biased on source data. It cannot deal with distant source and target data since the earlier layers are frozen (no learning capacity for additional detailed features). However, it can deal with task transfer better than the other approaches because of new lateral layers \cite{epp}.

The final goal of Artificial General Intelligence (AGI) and the AI community is to imitate human intelligence and the learning process. Human intelligence and wisdom are the results of lifelong learning and training. We build knowledge on top of previously learned knowledge in continual/progressive processes. E.g., we learn about simple shapes and objects (rectangles, circles, squares, etc.). As we grow using that fundamental knowledge, we learn more complex shapes and objects (cube, octagonal, airplane, car, etc.) and how to distinguish them in a different context \cite{e3}.

We start to learn fundamentals in our childhood through exemplary (perfected) and simplified examples. We started learning shapes and letters by practicing basic examples in pre-school and kindergarten. We add to our knowledge and skills during K-12 school, building on what we previously learned. As our knowledge grows, we learn about more complex and ambiguous problems in a continuous learning process. A vital point of this learning process is that we start with the exemplary (perfect) and straightforward fundamental knowledge to reach wisdom after long-life training \cite{e3}. Inspired by this learning process, we introduce two-step training of deep learning models in EXPANSE.

Training processes in machine learning and deep learning are usually based on the training dataset's mixed distribution (mostly normal distribution). Deep learning models are initialized by random weights and start extracting features from details to abstract (first layers towards output layer) \cite{ey} with no prior knowledge. An extreme analogy of this regular training process would be asking a child to detect cancer tissues after showing him/her a thousand images of cancer tissues.

We have almost no idea what features the model is extracting in this training format and if those are valid for similar yet slightly different datasets. In other words, the extracted features could be strongly related to the specific training samples and not based on any fundamental knowledge. For instance, detecting cancer tissues could have been done by the shape of the area and not based on discoloration or texture of tissue. This is why a trained model for detecting animals can be deceived by changing the context/background in the image while a human can detect it in different contexts.

EXPANSE introduces a new system of continual learning for deep learning inspired by the human education system and the current progressive learning technique. The first aspect of this proposed system is to change the process of training deep learning models based on the human learning process. The second is to expand the network while expanding the knowledge in the vertical dimension by adding new nodes on pre-trained layers. We also consider that we may need to expand the network horizontally for some transfers, similar to how progressive learning adds new layers to the model. In deep learning, earlier layers extract detailed features, and the lateral layers towards the output extract more abstract knowledge using the extracted detailed features \cite{ey}. In EXPANSE, we offer to expand layers by adding new nodes to them to increase the model learning capacity. We show that in this way, not only can the final model deal with the target data; it is still valid on source data, which opens a reliable path to reach continual learning.

\section{Related work, problem, and motivation }

In the process of training the DL model, the learning rate adjusts the size of the steps in the gradient descent on the error surface \cite{e37,e38}. It is a tricky parameter (hyper-parameter) since if it is too small, the model may never move enough on the error surface to find the local (global) minima. On the other hand, the model may jump over local minima basins and never find them if the step is too big. Numerous studies have been done on this topic, e.g., variable learning rate \cite{e37,e39}. However, we could not find any study that adjusts the learning rate based on the quality of training data in deep transfer learning.

Even though there are many successful studies in the deep transfer learning field, the methods are mostly tailored to a specific dataset and a task, and the success is mainly achieved based on trial and error. As mentioned in the introduction section, most existing deep transfer learning (DTL) methods are network/model-based approaches and can be categorized into three groups \cite{e2}.

The first category is based on finetuning a pre-trained model. In this case, a related pre-trained model, trained on source data, will be used and trained (finetuned) on the target data. The primary issue in this approach is known as catastrophic forgetting dilemma since the obtained knowledge can be partially or totally wiped out \cite{ec,ec2}, making it a foremost necessity for this method to use similar source and target data. Catastrophic forgetting happens because the weights throughout the model can drastically change in the process of finetuning on target data \cite{e2,ec2}. However, if the source and target data are close enough, there is a high chance of success \cite{ef0,ef1,ef2,ef3}.

The second category is mainly for deep CNN models. In this method, the pre-trained CNN layers are frozen for the target domain training process with a new header and lateral fully-connected layers \cite{e2}. Change of header helps adjust the input shape, and change to the fully-connected layers brings the possibility of changing the model objectives, while the frozen CNN layers do the feature extractions from details to abstract. This technique reduces the effect of catastrophic forgetting to some extent. However, the model is still strongly biased on source data and cannot adapt to target data. Success depends on whether the source and target data are tightly related. Therefore, to increase the success rate, it is a known practice to choose a profoundly robust pre-trained model on a massive dataset such as ImageNet \cite{ein} to ensure that source data is broad enough to contain target data to some extent. Such pre-trained models are available publicly \cite{ein}, which is why this method has been used extensively. \cite{efz0,efz1,efz2,efz3,efz4} are successful examples of using this method.

The third category of model-based approaches is progressive learning, which aims to address both catastrophic forgetting and adaptability to target data and task. In progressive learning, the whole pre-trained model is used frozen while a new layer(s) adds to the end of the model for the process of training on target data \cite{epp}. This approach can deal better with slightly less related source and target data. However, it still suffers from being biased on source data when the target has more features or different features than the source. Some successful examples of using this approach are \cite{ep1,ep2,ep3,ep4}.

\cite{egb} is an intriguing study of expanding the CNN model both horizontally and vertically in the process of finetuning. This is the closest related work to EXPANSE that we found. They try to evaluate and compare the effects of expanding the model in two ways, adding new layers vs. adding units on existing layers. Their results prove that expanding the model (or "growing" the model, as they call it) in either direction is helpful in the transfer of knowledge and obtaining better accuracy on target data. They conclude that adding units on some layers instead of adding new layers has a slight but consistent benefit in their experiments, which aligns with our EXPANSE assumptions even though their study's mindset and evaluation setup are not the same. However, they only explore finetuning without freezing any part of the pre-trained model, even theoretically. Also, their method is specific to CNN layer expansion and only on some lateral layers, which, as we mentioned, results in a biased pre-trained model without the capacity to learn any new detailed features (on earlier layers).

\section{EXPANSE}

EXPANSE's design is based on the human education system and is inspired by progressive learning \cite{epp}. The main objective is to improve deep transfer learning (DTL) by dividing the problem and the associated model into more straightforward and simple steps. Then, through a continual/progressive learning approach, gradually expand the model and training samples. Thus, we named our proposed system EXPANSE. \cite{e3}

The first aspect of EXPANSE is about dividing the training samples. We will add a limited number of perfected samples or select limited exemplary samples from the training dataset depending on the training data. We train the model first with the exemplary samples at a higher learning rate (LR), then finetune that model with the whole training data (containing exemplary samples as well) at a lower learning rate (LR). We call this approach two-step training. This technique aims to help the model extract more meaningful features and find a finer basin on the error surface with larger steps (higher learning rate) using a limited number of exemplary samples; then explore that area of the error surface to find the local minima with smaller steps (lower learning rate). This idea follows our learning process in schools. We first learn the fundamentals, then navigate uncertainties and more complex problems.   

The other aspect of EXPANSE is about model expansion in the process of deep transfer learning. Increasing the model's size in either direction increases the model's learning capacity. As we mentioned, the earlier layers in deep learning extract detailed features. And moving towards the output, the layers are responsible for extracting abstracts by the obtained detailed features \cite{ey}. Therefore, if the target domain contains more features than the source domain, adding layers towards the end of the model is not enough. The model cannot extract those detailed features on a frozen model or replace the previously gained knowledge on earlier layers. As we mentioned in the previous section, these two extremes are known as an overly biased pre-trained model with frozen layers or catastrophic forgetting dilemma in the case of finetuning. 

\begin{figure}[h]
  \centering
  \includegraphics[width=\textwidth]{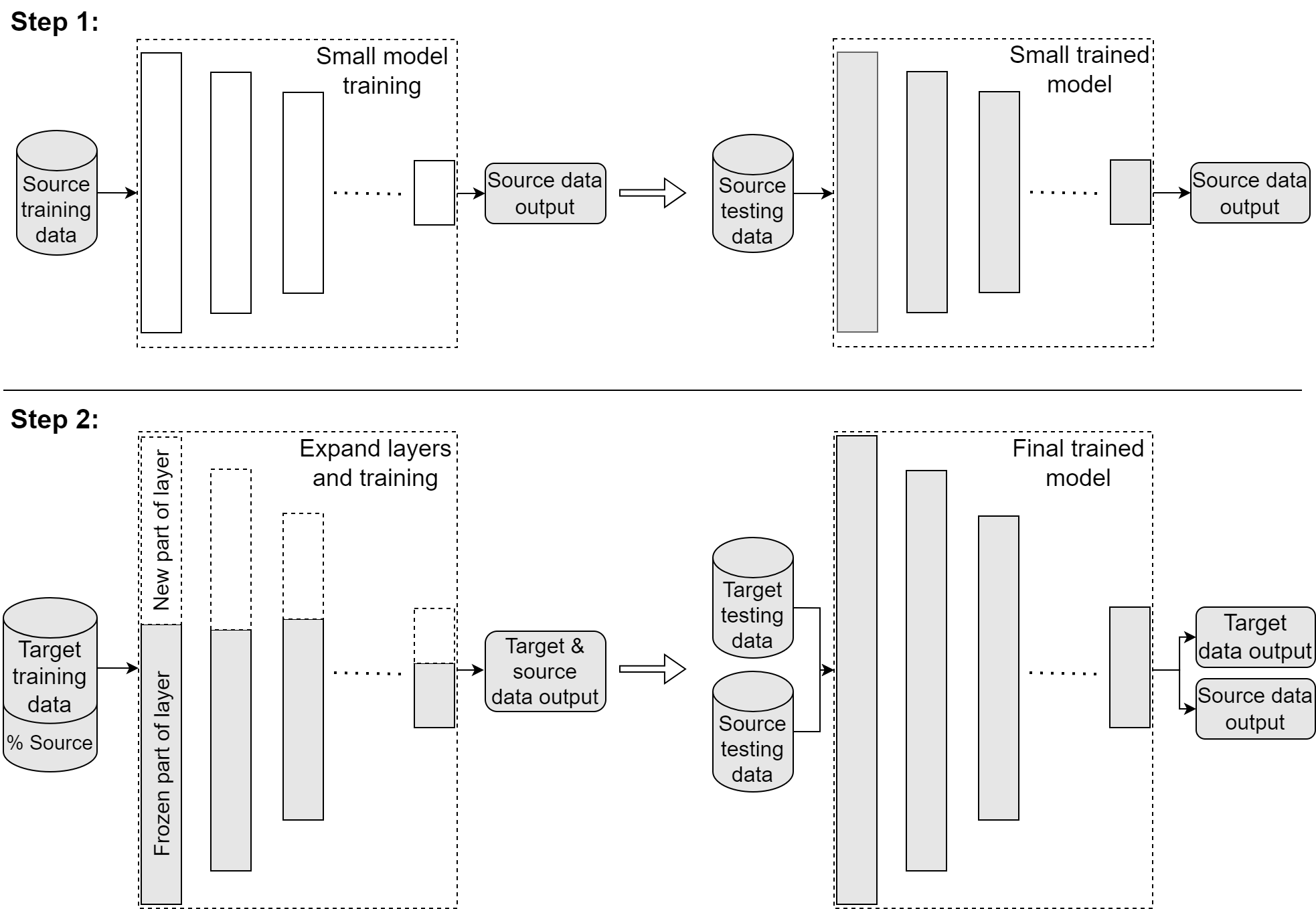}
  \caption{Model expansion illustration.}\label{fig2}
\end{figure}

To address this, in EXPANSE, we consider that the model can expand in both directions (new nodes on existing layers and new layers) while the pre-trained section of the model will be used frozen. The expansion of layers (adding new nodes) will increase the learning capacity of the network. The earlier layers can extract more detailed features, and lateral layers can extract more complex abstract knowledge, as well as the output layer. Adding the new lateral layer(s) becomes crucial when the model objective changes drastically towards more complex tasks in the process of deep transfer learning. Figure \ref{fig2} illustrates the vertical model expansion in EXPANSE. The continual learning in EXPANSE happens first because the expanded model stays valid for source data. The final model can also be again expanded for another step of training on new target data, and this process can be repeated.

Decision-making regarding model expansion size is similar to the deep learning model hyper-parameter adjustments. There is no exact formula, but through experience and analysis of data and tasks, an expert can narrow down the size of the model expansion and the parameters in a limited number of trials. In EXPANSE, the expansion size should be decided by common sense and experience based on the difference in tasks and training data in each step of deep transfer learning. A key to successful expansion is to consider that the earlier layers need expansion when the target data offers more detailed features, and the middle layers should expand when we expect the model abstract knowledge extraction to increase. 

The EXPANSE is a methodology to be applied to different problems and DL models, not a solution for a specific case. We are introducing a systematic continual learning approach to deep transfer learning. The goal is to improve the process of transfer learning in deep learning and open a door to a more sophisticated approach to continual learning to move the current AIs a step closer to AGI (human-level continual/progressive learning).

\section{Evaluation}

EXPANSE is a methodology applicable to machine learning models such as deep neural networks and deep CNN models. To evaluate such a methodology, we need to verify its practicality and performance with more general tasks and compare results to a similar model and task without the EXPANSE methodology applied. Such evaluation aims not to prove that this approach can outperform other techniques for a singular task; as long as the method is applicable and results are similar to a model without EXPANSE, we have achieved our goal. The goal is to open the door to continual learning possibility, not compete on a single task. However, our results proved the applicability and practicality of EXPANSE and showed unanticipated performance improvements.

One major limitation of the proposed system is the implementation of adding new nodes to frozen layers since it requires a structural change to the model's core. We need to be able to freeze part of a layer and let the new nodes be trainable on the target data on that layer. Today, none of the existing framework's libraries for DLs supports such an ability. Also, in the case of the CNN layer, this implementation is even more complicated; however, it might be dealt with by increasing the number of channels in that layer and only freezing the pre-trained channels in that layer. 

To evaluate EXPANSE, we use deep neural networks, and instead of freezing the pre-trained section, we finetune the expanded model. The Algorithm \ref{alg1} shows the EXPANSE algorithm, including this limitation. We consider dealing with this technical implementation issue of freezing part of a layer as future work. However, the results show the potential of this methodology even without freezing and demonstrate that being able to freeze the pre-trained section should increase the performance of EXPANSE in the future.

\begin{algorithm}[h]
\caption{EXPANSE}\label{alg1}
\begin{algorithmic}
\Require source data, target data, exemplary source data, exemplary target data
\State $Model A \gets First model$ {for source data (random initialized)}
\State $Model B \gets Train$  {A  on  exemplary  source data}
\State $Model C \gets Fine tune$  {B on exemplary source data + source data}
\State $Model D \gets Expand$  {C  with new nodes (random initialized) on layers}
\State $Model E \gets Freeze$  {the C part of model D}
\If{$E$ is done} {(future work)}
    \State $Model F \gets Train$  {E on exemplary target data}
    \State $Model Final \gets Finetune$  {F on exemplary target data + target data}
\ElsIf{$E$ is NOT done} {(current evaluation)}
    \State $Model Final \gets Finetune$  {D on target data + exemplary target data + exemplary source data + a portion of source data}
\EndIf
\State $return (Model Final)$
\end{algorithmic}
\end{algorithm}

We chose the MNIST dataset \cite{em,emk} and a deep neural network model similar to \cite{ems, emk} for our evaluation process. This dataset consists of 60,000 handwritten digits for the training set and 10,000 samples for testing. MNIST is a well-known dataset in the Machine Learning (ML) community to benchmark new methods \cite{emb}. Also, it is possible to create exemplary perfected data using printed digits for evaluating our proposed two-step training.

We consider the end model (after expansion) as the final model to have a benchmark for comparison. We train the final model from randomly initialized weights and consider the obtained results as the benchmark. We save our random initialized weight and start all the training with the same weights to make a fair comparison. Also, through all tests, we keep the hyper-parameters constant. All experiments are easily reproducible since we have used random seeding so that any run will end up with the same results. The experiments were done on Google Colab using python and TensorFlow (Keras) libraries \cite{egckp}, accessible at \cite{eexp}.

For this experiment, we made 180 perfected samples of ten printed digits using 18 fonts, and by simply duplicating the samples, we made a perfected dataset of 360 samples. Our final model consists of three (3) layers of 256, 128, and 10 nodes, using Relu activation function, Adam optimizer, and 10-fold cross-validation \cite{eafo,e1}. Our experiments' epoch number is limited since we use 10-fold cross-validation; one epoch means nine (9) times running through the training data.

In the following experiments, we first evaluate two-step training separately and then evaluate the EXPANSE methodology: model expansion and two step-training.

To verify the effectiveness of the two-step training method, we train the final model with the random initialized value with two-step training; first, train on perfect data and then add handwritten samples to compare the result with the benchmark. Moreover, we trained the final model with the randomly initialized weights by mixing the perfect data and the MNIST training dataset. 

Table \ref{table1} shows the obtained results from the two-step training evaluation. As listed in the first two rows (the benchmarks), after getting trained by 60,000 handwritten samples, the model can only be 57.22\% accurate in detecting printed digits in the best case. The expectation from such training is that the model should easily identify perfect samples, but the result aligns with our assumption that the usual training process of deep learning is not always meaningful or capable of building on fundamental knowledge. On the other hand, the model trained by only 360 (0.6\% of the size of MNIST training dataset) exemplary perfect data samples in less than 30 seconds can detect around 40\% of handwritten samples (row number 3). Row number 4 shows the results of finetuning the model from row number three on only the training dataset of MNIST, which shows the catastrophic forgetting dilemma since the accuracy of perfect data dropped. To avoid the catastrophic forgetting dilemma, we add the perfect samples to the target data (MNIST training data) for the finetuning process (row number 6). 

The accuracy of the two-step trained model implies that there existed a finer local minima on the error surface, which could work for both the perfected data and the handwritten data to improve the accuracy of handwritten test data. In other words, the model found a more meaningful basin on the error surface. Also, even adding the perfect samples to the MNIST training data for training the randomly initialized model in one step (row number 5) while improving the results is still not as accurate as two-step training (row number 6). This follows the logic behind the human education system: the most effective way of learning is to focus first on the basics and then add complexity. 

It is worth mentioning that increasing the number of epochs in these tests empowered the effectiveness of the two-step method. We limited the number of epochs to limit the training time for the purpose of reproducibility and the verification process. For instance, the same experiment with eight (8) epochs resulted in 98.33\% (two-steps training) vs. 98.07\% (traditional training) accuracy on the MNIST test dataset.

\begin{table}[h]
  \caption{Two-step training on MNIST}
  \label{table1}
  \centering
  \begin{tabular}{ p{0.02\linewidth}  p{0.52\linewidth}   p{0.11\linewidth}   p{0.1\linewidth}   p{0.09\linewidth} }
    \toprule
    \multicolumn{4}{r}{Accuracy on}                   \\
    \cmidrule(r){3-5}
    \#   & \centering Description    &  Exemplary data     &   MNIST train data &  MNIST test data \\
    \midrule
    1 & Random initialized on MNIST with LR=0.001 \& epoch=3 & 55.56\% & 99.93\% & 98.04\%  \\
    2 & Random initialized on MNIST with LR=0.002 \& epoch=3 & 57.22\% & 99.57\% & 97.83\%  \\
    3 & Random initialized on Perfect data with LR=0.01 \& epoch=8 & 100.00\% & 39.54\% & 41.24\%  \\
    4 & Fine-tune the pre-trained model on only MNIST with LR=0.002 \& epoch=3 & 68.33\% & 99.93\% & 97.97\%  \\
    5 & Random initialized on Mix data with LR=0.002 \& epoch=3 & 100.00\% & 99.71\% & 98.02\%  \\
    6 & (Two-steps training) Fine-tune the pre-trained model on Mix data with LR=0.002 \& epoch=3 & 100.00\% & 99.79\% & 98.06\%  \\
    \bottomrule
  \end{tabular}
\end{table}

To verify the applicability and effectiveness of model expansion along with two-step training, the EXPANSE system, we divided the MNIST training data and exemplary data into two sets of 0 to 4 digits and 5 to 9 digits. The MNIST training data consists of 30,596 samples of 0 to 4 and 29,404 samples of 5 to 9 digits. We reduced the final model size to three (3) layers of 150, 80, and 5 nodes for the first training step. First, we trained the small model using the 180 exemplary samples and then finetuned it with a mix of MNIST training data and exemplary samples for digits 0 to 4. For applying the model expansion, we updated the weights of the small portion of the random initialized final model (three (3) layers of 256, 128, and 10 nodes) with the weights of a finetuned small model. Then we trained that model using the mix of MNIST training data and exemplary samples of 5 to 9 with a part of the samples from 0 to 4. We mixed some samples of the previous step into the final step since we could not freeze the small section of the network, and we wanted to avoid catastrophic forgetting. 

Table \ref{table2} shows the obtained results at each step of this process. The first two rows are only for samples of 0 to 4 digits. As listed on row number 4, EXPANSE methodology has improved the same model's (size and configuration) accuracy from 98.04\% (Table \ref{table1}, row number 1) to 98.09\% on MNIST test data and from 55.56\% to 100\% on printed digits. This demonstrates the practicality and effectiveness of the EXPANSE continual learning system and shows that this methodology can improve the performance of the same model (size and configuration).

\begin{table}[h]
  \caption{Expanse on MNIST}
  \label{table2}
  \centering
  \begin{tabular}{ p{0.02\linewidth}  p{0.52\linewidth}   p{0.11\linewidth}   p{0.1\linewidth}   p{0.09\linewidth} }
    \toprule
    \multicolumn{4}{r}{Accuracy on}                   \\
    \cmidrule(r){3-5}
    \#   & \centering Description    &  Exemplary data     &   MNIST train data &  MNIST test data \\
    \midrule
    1 & Small model, random initialized on perfect (0 to 4) with LR=0.01 \& epoch=3 & 100.00\% & 64.62\% & NA  \\
    2 & Small model, fine-tune on mix data (0 to 4) with LR=0.002 \& epoch=3 & 100.00\% & 99.97\% & NA  \\
    3 & Final model (loaded weights from small model), fine-tune on mix data with LR=0.001 \& epoch=3 & 100.00\% & 99.89\% & 98.04\%  \\
    4 & Final model (loaded weights from small model), fine-tune on mix data with LR=0.002 \& epoch=3 & 100.00\% & 99.87\% & 98.09\%  \\
    \bottomrule
  \end{tabular}
\end{table}

A key point in this set of experiments is that the digits 0, 1, 2, 3, and 4 do not share the same visual features as 5, 6, 7, 8, and 9. Therefore, in this case, the source and target data are not closely related and can be considered distant datasets, and EXPANSE successfully handled it.

Further, we selected the best deep neural network (DNN) model without any pre-processing (e.g., distortion techniques) of training data from the list at \cite{em} to verify if, by applying EXPANSE, we can improve such a model. The selected model is listed as "3-layer NN, 500+300 HU, softmax, cross entropy, weight decay", with an accuracy of 98.47\%. We first implemented their model and obtained 98.46\% accuracy on MNIST test data, table \ref{table3} row number 2. Then, we applied EXPANSE to the same model, the first step with a smaller model of 300, 200, and 5 nodes and the second step of 500, 300, and 10 nodes. We obtained 98.52\% accuracy, table \ref{table3} row number 3. Not only do we improve the accuracy of the existing model on MNIST test data, but our model can also detect printed digits perfectly. It is worth mentioning that a pre-processing technique or use of CNN layers is necessary to gain accuracy above 99\% on MNIST \cite{emc}.

\begin{table}[h]
  \caption{Expanse on MNIST vs. best existing deep neural network (DNN) model}
  \label{table3}
  \centering
  \begin{tabular}{ p{0.02\linewidth}  p{0.52\linewidth}   p{0.11\linewidth}   p{0.1\linewidth}   p{0.09\linewidth} }
    \toprule
    \multicolumn{4}{r}{Accuracy on}                   \\
    \cmidrule(r){3-5}
    \#   & \centering Description    &  Exemplary data     &   MNIST train data &  MNIST test data \\
    \midrule
    1 & Random initialized on MNIST with LR=0.0009 \& epoch=5 & 58.33\% & 99.99\% & 98.42\%  \\
    2 & Random initialized on MNIST with LR=0.0009 \& epoch=7 & 53.89\% & 99.99\% & 98.46\%  \\
    3 & Final model (loaded weights from small model), fine-tune on mix data (66\% of source data) with LR=0.0009 \& epoch=5 & 100.00\% & 99.75\% & 98.52\%  \\
    4 & Final model (loaded weights from small model), fine-tune on mix data (10\% of source data) with LR=0.0009 \& epoch=5 & 100.00\% & 97.65\% & 96.55\%  \\
    \bottomrule
  \end{tabular}
\end{table}

In this experiment, we used two-thirds of the source data mixed with target data to avoid catastrophic forgetting for the second step. However, even with using only 10\% of source data mixed with target data for the second step of training, we achieved 96.55\% accuracy on the MNIST test dataset, table \ref{table3} row number 4. This very slight drop in accuracy shows that even without being able to freeze the small part of the trained model after the first step, the model is not dramatically losing the obtained knowledge from source data while getting trained on target data. In our opinion, the reduction of the catastrophic forgetting effect in EXPANSE is because vertically expanding the model provides a capacity to learn new features without replacing the previously obtained knowledge. Moreover, this implies that by being able to freeze the smaller section of the model in the second step (future work), we will reduce training time and possibly improve accuracy since we will not need to use any source data (or a limited number of samples) during the training on target data. 

Following the promising results on MNIST, we also applied EXPANSE to a simple model of a three-layer (500, 300, and 10 nodes) neural network on Fashion MNIST \cite{efm}. Fashion MNIST dataset "comprising of 28x28 grayscale images of 70,000 fashion products from 10 categories, with 7,000 images per category. The training set has 60,000 images and the test set has 10,000 images. Fashion-MNIST is intended to serve as a direct drop-in replacement for the original MNIST dataset for benchmarking machine learning algorithms, as it shares the same image size, data format and the structure of training and testing splits." \cite{efm}

In this case, instead of making perfected samples, we manually selected a limited number of exemplary samples from the training dataset (18 samples for each of the 10 categories) based on the simplicity and clarity of the samples. Like previous experiments, we started with a smaller model (300, 200, and 5 nodes) and the first five categories. Next, we expanded the model to 500, 300, and 10 nodes and trained on all categories. Again, we use a part of the first step's training samples in the second step to reduce the catastrophic forgetting effect. Also, similar to previous tests, we trained the randomly initialized final model on the F-MNIST training dataset as the benchmark to compare our results with it. 

\begin{table}[h]
  \caption{Expanse on Fashion MNIST}
  \label{table4}
  \centering
  \begin{tabular}{ p{0.02\linewidth}  p{0.52\linewidth}   p{0.11\linewidth}   p{0.1\linewidth}   p{0.1\linewidth} }
    \toprule
    \multicolumn{4}{r}{Accuracy on}                   \\
    \cmidrule(r){3-5}
    \#   & \centering Description    &  Exemplary data     &   F-MNIST train data &  F-MNIST test data \\
    \midrule
    1 & Random initialized on F-MNIST with LR=0.0001 \& epoch=6 & 96.67\% & 96.79\% & 88.82\%  \\
    2 & Random initialized on F-MNIST with LR=0.0001 \& epoch=7 & 98.89\% & 97.86\% & 89.24\%  \\
    3 & Final model (loaded weights from small model), fine-tune on mix data (70\% of source data) with LR=0.0001 \& epoch=6 & 100.00\% & 98.43\% & 89.42\%  \\
    4 & Final model (loaded weights from small model), fine-tune on mix data (70\% of source data) with LR=0.0001 \& epoch=7 & 100.00\% & 98.73\% & 89.43\%  \\
    5 & Final model (loaded weights from small model), fine-tune on mix data (10\% of source data) with LR=0.0001 \& epoch=6 & 100.00\% & 90.24\% & 84.39\%  \\
    \bottomrule
  \end{tabular}
\end{table}

The results on F-MNIST are consistent with our earlier MNIST experiments; both two-step training and model expansion achieved better accuracy as well, table \ref{table4} row 1 and 2 vs. 3 and 4. Even using a set of exemplary data from the same training set and applying two-step training improved the model accuracy. Moreover, similar to MNIST experiments reducing the reuse of source data in the target training process to 10\% did not drastically drop the accuracy, table \ref{table4} row number 5.

These results convey the capability of EXPANSE in continual learning, meaning that EXPANSE can be applied to a chain of problems, and the final model can answer all the problems and reuse the knowledge obtained at every stage of training. Moreover, the models from the intermediate steps of such a chain of training could be used later for different tasks. For example, a final task of image classification of interior objects could be done through steps of 1) simple shapes, 2) colors, 3) furniture, 4) building objects, 5) arts, 6) plants, etc. So, the model from step 4, building objects (door, window, stairs, etc.) could be separately used (trained on more data) for interior and exterior building object classification.

\section{Conclusion}

Inspired by the human education system, we introduce the two-step training method for deep learning models: training the model first with limited exemplary data (learning basics first), then finetuning that model with the training data (learning to deal with uncertainties and complexity). This method can be considered an educational supervised learning. Our evaluation demonstrated the effectiveness of using two-step training compared to the traditional training of deep learning models. Furthermore, model expansion, along with two-step training, the EXPANSE methodology, deals with catastrophic forgetting dilemma and overly biased models in deep transfer learning by increasing the whole model's learning capacity through vertical expansion of the model. Adding new nodes to the pre-trained layers increases the learning capacity of the model, even in the case of extracting detailed features for unrelated source and target data. Moreover, this approach opens a reliable path to continual learning in deep transfer learning since our final model is still valid on source data. Even with implementation limitations (freezing part of a layer), our evaluation demonstrates the high potential of the EXPANSE approach for a successful transfer in deep learning that can be applied to different DL models and tasks. 

\section*{Ethics}

This work does not have any direct societal impacts. However, scientific advancements introduce ethical issues that are unavoidable and often not easy to predict. In \cite{e1}, we discussed ethical issues stemming from AI development. Here we would like to emphasize our previous call to data scientists and the AI community to partner with other disciplines (e.g., social and behavioral sciences) to consider the effects of their creations on society, no matter how far into the future they may reach.

\bibliographystyle{unsrtnat}
{
\small
\bibliography{biblio}
}

\end{document}